\documentclass[a4paper,11pt]{article}

\usepackage[english]{babel}
\usepackage[utf8]{inputenc}
\usepackage{amsmath}
\usepackage{graphicx}
\usepackage[colorinlistoftodos]{todonotes}
\usepackage{siunitx}

\usepackage{graphicx}
\usepackage{booktabs}
\usepackage{caption}
\usepackage{url}
\usepackage{algorithm}
\usepackage{algpseudocode}
\usepackage{tabularx}
\usepackage{amsfonts}

\usepackage{multirow}
\usepackage{listings}
\usepackage{arydshln}

\usepackage{float}
\usepackage[colorinlistoftodos]{todonotes}
\usepackage[colorlinks=true, allcolors=blue]{hyperref}
\newcolumntype{P}[1]{>{\centering\arraybackslash}p{#1}}

\title{Improving Clinical Outcome Predictions Using Convolution over Medical Entities with Multimodal Learning}

\author{Batuhan Bardak, Mehmet Tan* \ \\
              Department of Computer Engineering \ \\
            TOBB University of Economics and Technology \ \\
             Ankara, Turkey \ \\
}

\date{}

\begin{document}
\maketitle

\begin{abstract}

Early prediction of mortality and length of stay(LOS) of a patient is vital for saving a patient's life and management of hospital resources. Availability of electronic health records(EHR) makes a huge impact on the healthcare domain and there has seen several works on predicting clinical problems. However, many studies did not benefit from the clinical notes because of the sparse, and high dimensional nature. In this work, we extract medical entities from clinical notes and use them as additional features besides time-series features to improve our predictions. We propose a convolution based multimodal architecture, which not only learns effectively combining medical entities and time-series ICU signals of patients, but also allows us to compare the effect of different embedding techniques such as Word2vec, FastText on medical entities. In the experiments, our proposed method robustly outperforms all other baseline models including different multimodal architectures for all clinical tasks. The code for the proposed method is available at \url{https://github.com/tanlab/ConvolutionMedicalNer}. 

\end{abstract}


\section{Introduction}
\label{sec:introduction}

Electronic Health Record (EHR) data collected from patients who have been admitted into hospitals or intensive care units (ICU) offer a detailed overview of patients consisting of but not limited to demographics, insurance, laboratory test results and medical notes. With the EHR data becoming available for researchers, there has been increasing interest in using it with deep learning algorithms. Besides rapid progress in deep learning area, after Medical Information Mart for Intensive Care(MIMIC-III)~\cite{johnson2016mimic}, today's most popular public EHR database, was released, numerous studies have achieved successful results using this data set and deep learning models to predict different clinical outcomes~\cite{ghassemi2017predicting,mcdermott2018semi,barton2019evaluation}. 

Understanding the health condition of the patient by observing the clinical measurements, laboratory test results and predicting the condition of patients during their ICU stay is a vital problem. In this paper, we focus on two different common risk prediction tasks, mortality (in-hospital \& in-ICU) and length of ICU stay (LOS). Both are very important clinical outcomes for determining treatment methods, planning hospital resources and ultimately saving lives. Previous studies primarily focused on predicting clinical events using only the structured data of patient such as historical patient diagnosis (ICD codes)~\cite{choi2016doctor, choi2016retain}, lab results and patient ICU measurements~\cite{caballero2015dynamically, song2018attend, suresh2018learning} and did not benefit from the unstructured data in EHR. The EHR data which consists of clinical notes written by doctors, nurses, or radiology, discharge notes and many other sources, contains quite detailed information about patients, projecting the knowledge and inference of doctors and even critical details about patient health status for many cases. As per the importance of the clinical notes, researchers want to take advantage of the rich content in clinical notes. Moreover, the recent developments in Natural Language Processing (NLP), there has been increasing interest in using clinical notes to make clinical model predictions~\cite{mullenbach2018explainable, boag2018s}. Although it may be possible to leverage clinical notes to make more accurate predictions, these notes may consist of long written free-text with an unusual grammatical structure and may contain redundant information. As it may be hard to process raw clinical notes, because of their high-dimensional and sparse nature, extracting medical entities is required to unlock the medical information trapped in the clinical notes and to feed them into prediction models.

Named Entity Recognition (NER) is a fundamental task in NLP that focuses on information extraction aiming to extract entities in a text and classify them into predefined classes. These classes can be locations, people, or organizations in general NER algorithms~\cite{manning2014stanford,honnibal-johnson:2015:EMNLP}. There can be various NER models for different domains like cybersecurity~\cite{gasmi2018lstm} or medicine~\cite{kormilitzin2020med7}. Recently, several deep learning algorithms were applied to clinical texts to train clinical named entity recognition models. These clinical NER models generally try to extract medical information such as disease, drugs, dosage, frequency.

In this paper, we argue that the integration of structured data in EHR and medical entities positively affects the prediction of mortality and LOS. We also investigate the effect of different word representations such as Word2Vec\cite{mikolov2013efficient}, FastText\cite{joulin2016bag}, and concatenation of both representations on medical entities. To evaluate the success of our proposed multimodal architecture, we first train models separately with structured and medical entity features. Then we apply multimodal approach and use these features together in several ways to show the effectiveness of the proposed network. The results indicate a promising increase in performance on mortality and LOS tasks when the medical entities are used with structured data in a multimodal approach.

In the next section, we summarize the similar studies that work on clinical domain especially predict mortality and length of stay at ICU. Following that, we discuss our data set, problem definitions, and deep learning models used in this study. Finally, we report experimental results and conclude the paper by our findings and conclusion.

\begin{table*}[!t]
\centering
\resizebox{\textwidth}{!}{\begin{tabular}{l c c c}
\hline
                                                 & \# of Patient & \# of hospital admission & \# of ICU admission \\ \hline
MIMIC-III ( $>$ 15 years old)                                       & 38,597         & 49,785                    & 53,423              \\ 
MIMIC-Extract                                    & 34,472         & 34,472                    & 34,472               \\ 
MIMIC-Extract (at least 24+6 (gap) hours patient) & 23,937         & 23,937                    & 23,937               \\ \hline
\textbf{Final Cohort} (After clinical note elimination)                        & 21,080         & 21,080                    & 21,080               \\ \hline
\end{tabular}}
\caption{Summary statistics of the original MIMIC-III dataset, and the final cohort that is used in this study.}
\label{table:summary-datasets}
\end{table*}

\section{Related Work}
\label{sec:related-work}

With the rapid development of deep learning algorithms in the last decade, the number of deep learning models increased substantially for various clinical predictions. Several studies have explored EHRs to solve clinical problems, e.g., ~\cite{lipton2015learning} used 13 different vital measurements to classify 128 diagnoses using Long Short Term Memory (LSTM) and DoctorAI~\cite{choi2016doctor} used Gated Recurrent Unit (GRU) to predict multi-label diagnosis for the next visit.~\cite{choi2017using} proposed early heart failure detection using Recurrent Neural Networks (RNNs). 
Forecasting the LOS and mortality have been a popular clinical problem for healthcare researchers in recent years. In earlier studies~\cite{kim2011comparison, dybowski1996prediction, celi2012database} on mortality prediction, hand crafted features are selected and used simple machine learning models like logistic regression with different severity scores such as APACHE~\cite{knaus1981apache}, SAPS-II~\cite{le1993new}, and SOFA~\cite{vincent1996sofa}. Nowadays with the progress on deep learning, different architectures have been applied on EHR data to predict this kind of problems.~\cite{awad2017early} used ensemble learning to make an early mortality prediction and~\cite{sadeghi2018early} proposed a method to predict mortality using 12 features extracted from the vital signals in the first hour of ICU admission. Darabi et al.~\cite{darabi2018forecasting} used Convolutional neural network to predict long-term mortality risk on the MIMIC-III dataset. More recent work\cite{song2018attend} includes attention to their deep learning model to improve models' success.
Another work~\cite{yakovlev2018prediction} try to predict LOS for acute coronary syndrome patients. There is a comprehensive survey on mortality prediction and LOS~\cite{awad2017patient}. Despite these studies and developments, one of the major problems that the healthcare researchers experienced, the researches on the literature are short of standardized preprocessing steps such as unit conversion, handling outlier and missing values, and transforming raw structured data into usable hourly time series data. In order to solve this problem, ~\cite{harutyunyan2019multitask, wang2020mimic, purushotham2018benchmarking} carried out a comprehensive benchmark on MIMIC-III for various tasks such as mortality, LOS, readmission, phenotyping and make their code publicly available. Purushotham et. al. ~\cite{purushotham2018benchmarking} extracts 17 features from the MIMIC-III and works on hospital mortality, LOS and ICD-9 code group predictions. They compared their proposed super learner method with feedforward and recurrent neural network.~\cite{harutyunyan2019multitask} is another research that benchmarked their results on the MIMIC-III. They used multi-task learning approaches to predict four clinical prediction tasks such as risk of mortality, LOS, detecting physiologic decline, and phenotype classification. MIMIC-Extract~\cite{wang2020mimic} is the most recent work which is an open source pipeline for transforming MIMIC-III data into directly usable features. Their pipeline first transforms the raw vital sign and laboratory data into hourly time series and then apply some preprocessing steps such as unit conversion, outlier handling, imputing missing data. In this study, to increase reproducibility, we used MIMIC-Extract pipeline to featurize MIMIC-III data. 

We also use medical entities which are extracted from clinical notes to improve our model predictions. Clinical natural language processing and information extraction has been widely studied in recent years on clinical notes.~\cite{si2019deep, liu2018deep} proposed a deep learning based multi-task learning to make clinical predictions from clinical notes. 
~\cite{boag2018s} compared different embedding approaches such as Bag of Words (BoW), Word2Vec and LSTM on clinical note representation by evaluating the prediction performance on diagnosis prediction and mortality risk estimation. More recently, transformer-based architectures such as BERT~\cite{devlin2018bert}, XLNET~\cite{yang2019xlnet} gave state-of-the-art performance on different NLP tasks. These models are pre-trained on medical data, which is then fine-tuned on clinical text~\cite{huang2019clinicalbert, alsentzer2019publicly}. However, clinicians generally use medical jargon and shorthands when they take these clinical notes which makes hard to process directly. There are a number of studies in the field of clinical NLP which try to extract medical entities in clinical notes~\cite{zhu2018clinical, bhatia2019comprehend, fraser2019extracting}. In this work, we use med7~\cite{kormilitzin2020med7} which is developed for free-text electronic health record. Then, we combine these medical entities with structured data to benefit from multimodal approach. For a detailed overview on deep learning for natural language processing in the clinical domain, readers can refer to~\cite{wu2020deep}.

Multimodal learning is a key research area that uses multiple sources to predict unique tasks~\cite{ngiam2011multimodal}. This approach has shown success in image captioning tasks~\cite{karpathy2014deep}, visual question answering~\cite{ilievski2017multimodal} and speech recognition~\cite{mroueh2015deep}. In the healthcare research domain, \cite{khadanga2019using} combines unstructured clinical notes and structural time-series data for predicting in-hospital mortality, decompensation, and LOS. Similarly, \cite{shukla2020integrating} made unified mortality prediction and try to explore how physiological time series data and clinical notes can be integrated. The study by Jin. et al\cite{jin2018improving} is the closest to our work in terms of motivation. They made hospital mortality prediction by combining clinical notes and time series data. Clinical notes are represented with Doc2VecC~\cite{chen2017efficient} algorithm in two different ways. First, they directly combine clinical notes with time series data, second, they use neural network based clinical NER service to extract five types of medical entities and identify negated entities from clinical notes. After this pre-processing, they use the same representation with the first model and reported a 2\% increase in the Area Under ther Curve (AUC).

The difference of our paper from ~\cite{jin2018improving} and the main contributions of this work can be summarized as follows.

\begin{itemize}
    \item We work with four different clinical outcome such as in-hospital mortality, in-ICU mortality, LOS$>$3 and, LOS$>$7 rather than just in-hospital mortality.
    \item We compare different types of word embedding methods (Word2Vec, FastText, Concatenation), and discuss the effect these methods on medical entities.
    \item We propose a convolutional based deep learning model for combining clinical NER features with time series ICU features. We compare our proposed model with several benchmarks. 
\end{itemize}

\section{Materials and Methods}
\label{sec:methodology}

In this section, we begin by describing our dataset. The details of baselines and clinical NER model are explained next and finally we propose our multimodal deep learning models.

\subsection{Data}
\label{subsec:dataset}

We use the publicly-available MIMIC-III dataset which contains de-identified EHR data of 58,976 unique hospital admissions, 61,532 ICU admissions from 46,520 patients in the ICU of the Beth Isreal Deaconess Medical Center between 2001 and 2012. We use MIMIC-Extract~\cite{wang2020mimic}, an open source data extraction pipeline, to extract structured time series features in MIMIC-III. MIMIC-Extract mainly focuses on the patient's first ICU visit with some patient inclusion criteria. They eliminate data from patients younger than 15 years old and where the LOS are not between 12 hours and 10 days. This pipeline produces a cohort of 34,472 patients and 104 clinically aggregated time-series variables. In all of our experiments, we use the first 24 hours of patient's data after ICU admission and only consider the patients with at least 30 hours of present data like MIMIC-Extract. In our multimodal approach we combine medical entities with time-series variables. Before applying the clinical NER model on notes, we drop discharge summaries to avoid any information leak. Furthermore, we drop all clinical notes the chart time of which do not exist. After these steps, we drop all patients who do not have any clinical notes in 24 hours. The preprocessing on clinical notes are made similar to~\cite{khadanga2019using}. In the train-test split, for all clinical tasks, we split the data based on class distribution with 70\%/10\%/20\% ratio. Statistics of the final cohort and the others are summarized in Table~\ref{table:summary-datasets}.

\noindent\textbf{Problem Definition.} We mainly focus on two vital clinical prediction tasks, mortality(in-hospital \& in-ICU) and LOS($>3$ \& $>7$) at ICU. We use the same definitions of the benchmark tasks defined by MIMIC-Extract as the following four binary classification tasks. The explanation of these tasks and the class distributions are as follows:

\begin{enumerate}
  \item \textbf{In-hospital mortality}: Patient who dies during hospital stay after ICU admission (Significantly imbalanced, \%10.5). 
  \item \textbf{In-ICU mortality}: Patient who dies during ICU stay after ICU admission (Significantly imbalanced, \%7). 
  \item \textbf{Length-of-stay $>$ 3}: Patient who stays in the ICU longer than 3 days (Slight imbalanced, \%43.2).
  \item \textbf{Length-of-stay $>$ 7}: Patient who stays in the ICU longer than 7 days (Significantly imbalanced, \%7.9).
\end{enumerate}

\subsection{Baseline Models}
\label{subsec:baseline}
 In this subsection, we discuss our time-series baseline modal that we evaluate on each of our four benchmark tasks. Further, we explain clinical NER model, embedding approaches to represent medical entities and the multimodal baselines used in this study .

\setlength{\tabcolsep}{2pt}
\renewcommand{\arraystretch}{2}
\begin{table}[!h]
\centering
\begin{tabular}{l| c c c}
\hline
\textbf{Medical Entity} &\textbf{Total Count} & \textbf{Unique Count} & \textbf{Example} \\ \hline
\multirow{1}{*}{Drug} & 744778 & 18268 & Magnesium\\  \hline
\multirow{1}{*}{Strength} &  156486 & 10749 & 400mg/5ml\\  \hline
\multirow{1}{*}{Form}  & 40885 & 597 & suspension\\  \hline
\multirow{1}{*}{Route}  & 207876 & 1193 & PO\\  \hline
\multirow{1}{*}{Dosage}  & 126756 & 7239 & 30ml\\  \hline
\multirow{1}{*}{Frequency}  & 71285 & 3344 & bid\\  \hline
\multirow{1}{*}{Duration}  & 5939 & 1185 & next 5 days\\  \hline
\end{tabular}
\caption{The first column shows the type of medical entity, the second columns shows the total number of related entity found in clinical notes, and the third column shows the number of unique entity number. The last column shows the output of med7 for example sentence given from clinical notes.}
\label{tab:nerfrequency}
\end{table}

\subsubsection{Time Series Model}
\label{subsubsec:time-series-model}

We employ both Long Short Term Memory (LSTM)~\cite{hochreiter1997long} and Gated Recurrent Units (GRU)~\cite{chung2014empirical} networks to capture the temporal information between the patient features. As a result of time-series baseline experiments, GRU has shown a better AUC and AUPRC performance than LSTM up to $\%0.5$ - $\%1$, while using a simpler architecture. Therefore, we use GRU for all of the multimodal architectures. In general, GRU cell has two gates, a reset gate $r$ and an update gate $z$. With these gates, GRU can handle the vanishing gradient problem. 

We can iterate the mathematical formulation of GRU modal as follows:

$$
z_t  = \sigma(W_z x_t + U_z h_{t-1} + b_z) \\
$$
$$
r_t  = \sigma(W_r x_t + U_r h_{t-1} + b_r) \\ 
$$
$$
\hat{h}_t = \tanh(w_h x_t + r_t \circ U_h h_{i-t} + b_h) \\
$$
$$
h_t = z_t \circ h_{t-1} + (1 - z_t) \circ \hat{h}_t
$$
$$
\hat{prediction} = \text{sigmoid}(W_h h_{t}+b_h) \\
$$

where $z_t$ and $r_t$ respectively represent the update gate and the reset gate, $\hat{h}_t$ the candidate activation unit, $h_t$ the current activation, and $\circ$ represents element-wise multiplication. For predicting the mortality and LOS, a sigmoid classifier is stacked on top of the one layer GRU with 256 hidden units.

\begin{figure*}[t!]
\centering
\includegraphics[width=150mm,scale=1]{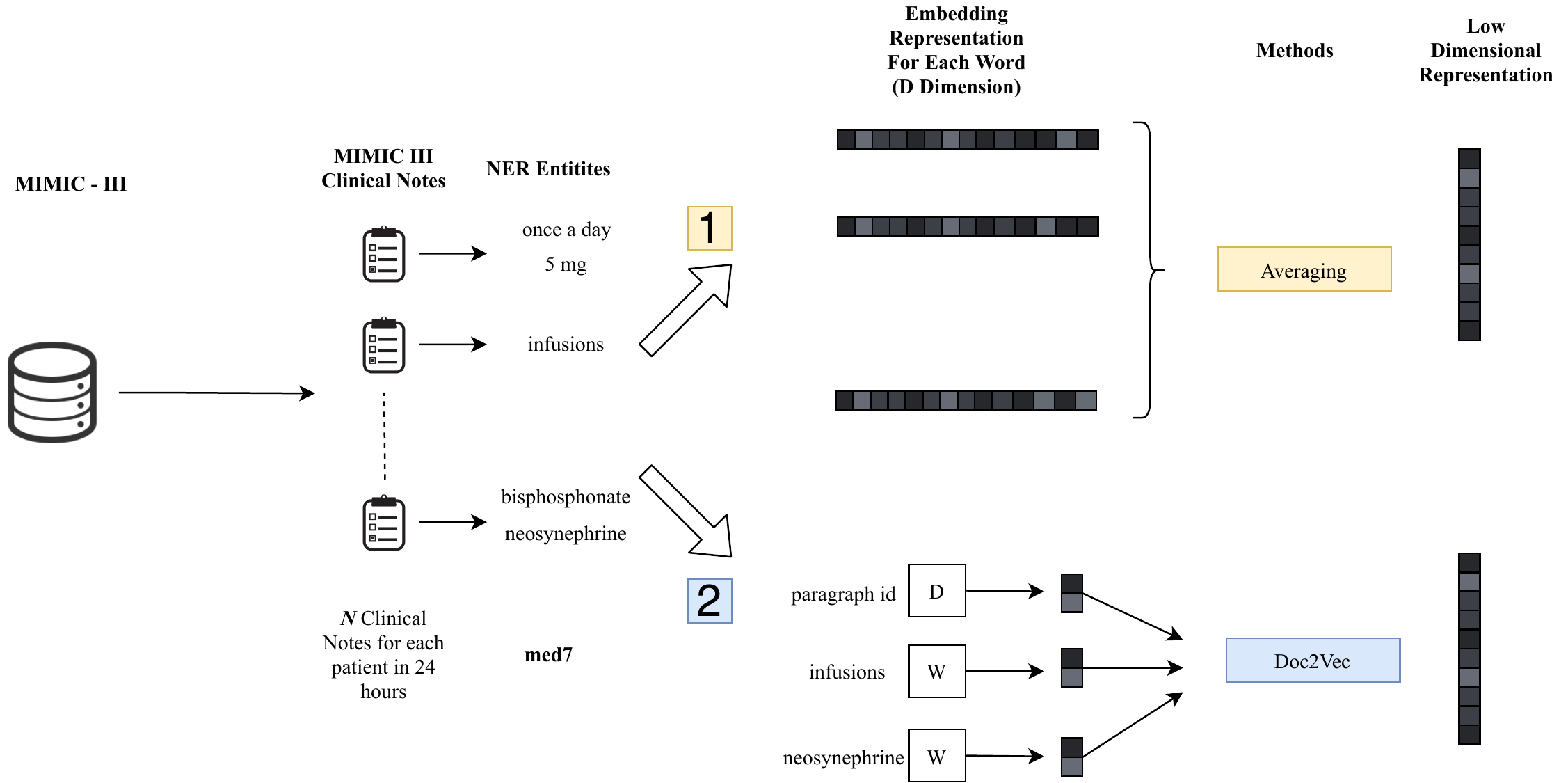}
\caption{Methodology for learning medical entity vectors. (1) The medical entities that are extracted from clinical notes are embedded into continuous word vectors. Then, we take the mean of these learned entity representations. (2) The words are removed from clinical notes if they are not belong to any medical entity category. Then, we train Doc2Vec on the preprocessed clinical notes to learn low dimensional representation of medical entities.}
\label{fig:1}
\end{figure*}

\subsubsection{Multimodal Approaches}
\label{subsubsec:multimodal-approach} 

In this work, besides time series features, we also use information from clinical notes to improve clinical task prediction performance. Instead of working directly with clinical notes, we first aim to extract medical related keywords. Recently, there are some notable works in the clinical domain that made their pre-trained clinical NER models publicly available~\cite{neumann2019scispacy, mulyar372tac, kormilitzin2020med7}. We use a pre-trained clinical NER model, med7~\cite{kormilitzin2020med7}, which uses the same dataset that we use in our experiments, MIMIC-III. This clinical NER model extracts seven different named entities such as 'Drug', 'Strength', 'Duration', 'Route', 'Form', 'Dosage',  'Frequency'. To represent the patient's medical entities we try two different embedding methods, word embedding and document embedding. First, we use three different word embedding algorithms to represent the each clinical NER model outputs and compare their performance. Second, we use Doc2Vec~\cite{le2014distributed} algorithm to represent the whole documents consisting of medical entities. The detailed schema of these two approaches are shown in Figure~\ref{fig:1} and the statistics of the extracted medical entities by med7 in MIMIC-III dataset for selected patients are shown in Table~\ref{tab:nerfrequency}.

\noindent\textbf{Word Embeddings.} Different word embedding methods might capture various semantic features on the same word. In our experiments, to understand this variety, we compare the performance of Word2Vec, FastText and the concatenation of Word2Vec \& FastText embeddings. Word2Vec~\cite{mikolov2013efficient} is a two-layer neural network that learns the representations of words in the given text with two ways: as a continuous bag-of-words (CBOW) and as a skip-gram. FastText~\cite{joulin2016bag} is an extension of the skip-gram model implemented by Facebook's AI Research (FAIR) lab which can handle out-of-vocabulary (OOV) words, and can learn better representations for rare words using several n-grams for words. We use pre-trained word2vec ($w_i \in \mathbb{R}^{100}$) and fastText embeddings ($f_i \in \mathbb{R}^{100}$) which was trained on 2.8 billion words from MIMIC-III clinical notes as shown in ~\cite{huang2019clinicalbert}. Lastly, we design an experimental embedding approach which concatanates the Word2Vec and FastText representations horizontally ($c_i \in \mathbb{R}^{200}$). When the Word2Vec embedding does not exist for a given word, we make zero padding in this setting.

\begin{figure*}[t!]
\centering
\includegraphics[width=165mm,scale=1]{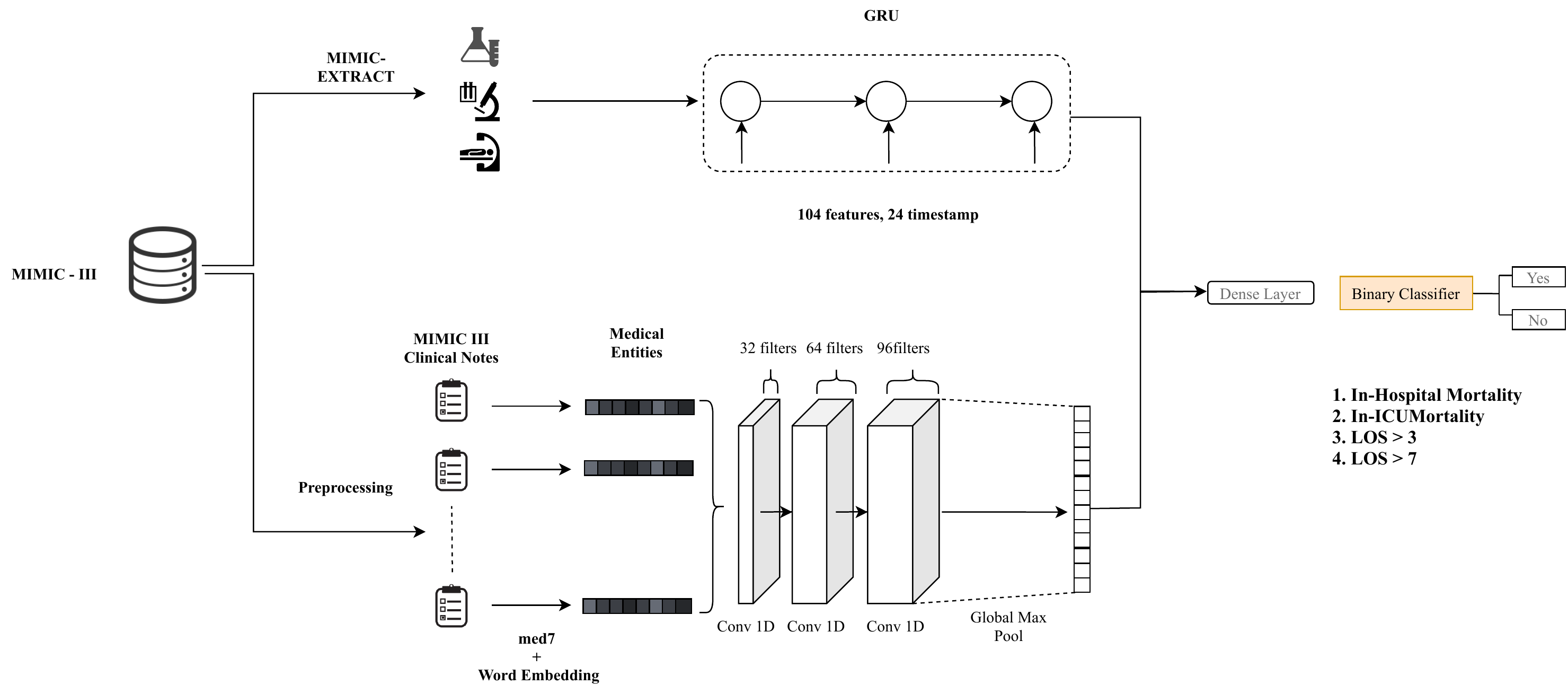}
\caption{Overview of Proposed multimodel architecture for predicting the In-Hospital Mortality, In-ICU Mortality, LOS $>$ 3, and LOS $>$ 7. To extract timeseries features, we use MIMIC-EXTRACT pipeline and fed these features through GRU. We also preprocess the clinical notes and use med7 to extract medical entities. 1D CNN is applied to extract features from medical entity representations. In the final layer, we concatenate features that extracted from timeseries and medical entities and fed through fully connected layer to predict 4 different binary clinical tasks.}
\label{fig:2}
\end{figure*}

\noindent\textbf{Document Embeddings.} Doc2Vec is an extention of Word2Vec model to learn document-level embeddings instead of word level. Before learning document level representations, we combine the first 24 hours of patient's clinical notes and apply clinical NER algorithm to keep only medical related keywords in the clinical notes. When training Doc2Vec, we use context window size of 5 words. This algorithm produces the fixed-length feature vector ($d_i \in \mathbb{R}^{100}$) for each patient.

\noindent We present two different baseline multimodal approaches with word and document embeddings that combine time-series data and medical entities.

\noindent\textbf{Multimodal with Average Representation.} This modal takes the average of all medical entities associated with a patient. For each patient, there are $N$ clinical notes and we extract $K$ medical entities from these $N$ clinical notes. Each medical entity is represented by word embeddings which is explained in Word Embeddings section. We sum $n$-dimensional $K$ clinical entities representation component wise and then divide this by $K$. We use two different input types to train our model. Time series data is processed through one layer GRU layer with 256 hidden units as explained in Section~\ref{subsubsec:time-series-model}. Averaged representations of medical entities are combined with time-series feature maps that are learned via GRU. In the end, these merged feature representations are fed into fully connected layer with 256 neurons, and a sigmoid classifier is added to the model.

\noindent\textbf{Multimodal with Doc2Vec Representation.} In this multimodal approach, instead of averaging medical entities, we apply Doc2Vec algorithm to obtain the fixed-length feature vector. First, we concatenate $N$ clinical notes for each patient and discard keywords from these notes if the keyword is not a medical entity. Then we apply the Doc2Vec algorithm to learn a low level representation from notes for each patient. After the learning fixed-length feature vector, we use the same architecture as average embedding approach.

\subsection{Proposed Model}
\label{subsection:proposed-model}

Figure~\ref{fig:2} describes the proposed multimodal approach which takes the advantage of 1D convolutional layers as a feature extractor on medical entities. Applying 1D Convolutional Neural Networks(CNN) on text learns the combination of adjacent words and shows successful results for various NLP problems~\cite{kim2014convolutional}. In our model, $K$ medical entities were extracted from $N$ clinical notes from each patient. These $K$ medical entities are first represented as a sequence of word embeddings with different word representation techniques such as Word2vec, FastText, and a combination of them. These entities $e_{i} \in \mathbb{R}^{d}$ are combined vertically and each patient is represented by a matrix $M \in \mathbb{R}^{k*d}$ where rows are filled with medical entity representations. This patient clinical NER entity matrix (padded where necessary) is represented as: 

\begin{equation}
\mathbf{e}_{1:k} = \mathbf{e}_1 \otimes \mathbf{e}_2 \otimes \ldots \otimes \mathbf{e}_k
\end{equation}

where $\otimes$ is the concatenation operator and $e$ refers to the representation of the medical entity and $k$ is the number of entity. We use a 1D-CNN model similar~\cite{ozturk2018deepdta} to extract features from medical entities. We stack three consecutive 1D convolutional layers with filter size 32, 64, and 96. The kernel size is same for three convolutional layer. The output of the last convolutional layer is followed by the max-pooling layer. The final features of the max-pooling layers are concatenated with the features from one layer GRU with 256 hidden units and fed through one fully-connected layer with 512 hidden units. 

\section{Experimental Results}
\label{sec:results}

\renewcommand{\arraystretch}{1.5}
\setlength{\tabcolsep}{10pt}

\begin{table*}[!t]
\centering
\resizebox{\textwidth}{!}{\begin{tabular}{l c c c c c}
\hline
Task & Baseline Modal & Embedding & AUROC & AUPRC & F1\\ \hline
\multirow{5}{*}{\textbf{In-Hospital Mortality}} 

& GRU & - & 85.04 $\pm$ 0.004 & 52.15 $\pm$ 0.009 & 42.29 $\pm$ 0.016\\
\cline{2-6}
 & Doc2Vec Multimodal & Doc2Vec & 85.96 $\pm$ 0.002 & 54.17 $\pm$ 0.004 & \textbf{46.60} $\pm$ 0.016\\
\cline{2-6}
 &                  & Word2Vec &\textbf{ 86.42} $\pm$ 0.004 & 54.22 $\pm$ 0.008 & 45.42 $\pm$ 0.013\\
 & Averaged Multimodal         &  FastText & 86.09 $\pm$ 0.004 & \textbf{54.47 }$\pm$ 0.007 & 45.50 $\pm$ 0.010\\
 &                  & Concat & 85.98 $\pm$ 0.002 & 54.19 $\pm$ 0.008 & 45.66 $\pm$ 0.021\\

\hline
\hline
\multirow{5}{*}{\textbf{In-ICU Mortality}} 
& GRU & - & 86.32 $\pm$ 0.004 & 46.51 $\pm$ 0.011 & 36.30 $\pm$ 0.026\\
\cline{2-6}
& Doc2Vec Multimodal & Doc2Vec & 86.80 $\pm$ 0.002 & 48.22 $\pm$ 0.006 & 41.95 $\pm$ 0.017\\
\cline{2-6}
 &                  & Word2Vec & \textbf{87.17} $\pm$ 0.002 & \textbf{48.47} $\pm$ 0.006 & 42.30 $\pm$ 0.021\\
& Averaged Multimodal                 &  FastText & 87.14 $\pm$ 0.003 & 48.36 $\pm$ 0.006 & \textbf{42.91} $\pm$ 0.014\\
 &                  &  Concat & 86.90 $\pm$ 0.004 & 48.28 $\pm$ 0.007 & 40.76 $\pm$ 0.022\\
\cline{2-6}

\hline
\hline
\multirow{5}{*}{\textbf{LOS $>$ 3 Days}} 

& GRU & - & 67.40 $\pm$ 0.003 & 60.17 $\pm$ 0.005 & 53.36 $\pm$ 0.016\\
\cline{2-6}
& Doc2Vec Multimodal & Doc2Vec & \textbf{68.90} $\pm$ 0.002 & \textbf{61.88} $\pm$ 0.002 & 54.32 $\pm$ 0.008\\
\cline{2-6}
 &                  &  Word2Vec & 68.63 $\pm$ 0.003 & 61.81 $\pm$ 0.003 & 54.19 $\pm$ 0.012\\
 & Averaged Multimodal &  FastText & 68.55 $\pm$ 0.003 & 61.59 $\pm$ 0.003 & 54.46 $\pm$ 0.012\\
 &                  &  Concat & 68.61 $\pm$ 0.003 & 61.69 $\pm$ 0.003 & \textbf{54.70} $\pm$ 0.009\\

\hline
\hline
\multirow{5}{*}{\textbf{LOS $>$ 7 Days}} 

& GRU & - & 70.54 $\pm$ 0.004 & 16.35 $\pm$ 0.006 & \textbf{2.33} $\pm$ 0.012\\

\cline{2-6}
& Doc2Vec Multimodal & Doc2Vec & \textbf{71.63} $\pm$ 0.005 & 17.22 $\pm$ 0.004 & 1.50 $\pm$ 0.007\\
\cline{2-6}

 &                  &  Word2Vec & 71.59 $\pm$ 0.005 & \textbf{17.91} $\pm$ 0.006 & 1.35 $\pm$ 0.008\\
 & Averaged Multimodal &  FastText & 71.31 $\pm$ 0.008 & 17.57 $\pm$ 0.007 & 1.02 $\pm$ 0.008\\
 &                  &  Concat & 71.59 $\pm$ 0.007 & 17.67 $\pm$ 0.007 & 1.37 $\pm$ 0.013\\

\hline
\hline
\end{tabular}}
\caption{Performance comparison of baseline methods. For all four clinical tasks, we report both AUC, AUPRC and F1 scores and the standard deviations.}
\label{tab:baseline-scores}
\end{table*}

In this section, we report the results of our baseline and multimodel experiments, the metrics we used for the evaluation and details about our development platform. 

\subsection{Setting}
\label{subsec:setting}

\noindent\textbf{Training.} For all tasks, we use the patient's first 24 hours ICU measurements. For multimodal architectures, we use 0.2 dropout rate at the end of the fully connected layer. A ReLU activation function is used for nonlinearity and $L_2$ norm for sparsity regularization is selected with the 0.01 scale factor. For the optimization, we use ADAM~\cite{kingma2014adam} algorithm with a learning rate of 0.001. All models are trained to minimize the binary crossentropy loss and we independently tune the hyperparameters - number of hidden layers, hidden units, convolutional filters, filter-size, learning rate, dropout rates and regularization parameters on the validation set. Each model is trained for 50 epochs and early stopping is used on the validation loss. We train each model 10 times with different initialization seed and report the average performance.

\noindent{\textbf{Evaluation metrics.}} The clinical problems that we work on suffer from class imbalance problem. We use three different metrics which are Area Under the Receiver Operating Characteristics (AUROC), Area Under Precision-Recall (AUPRC) and F1. AUROC is a popular robust metric for imbalanced datasets~\cite{davis2006relationship}. The second metric AUPRC does not include the true negatives in calculation and this approach makes it useful for data with many true negatives as our dataset. F1 is the final metric which calculates the harmonic mean of precision and recall.

\noindent{\textbf{Implementation Details.}} The aforementioned deep learning algorithms are implemented using Keras~\cite{chollet2015}, which runs Tensorflow~\cite{tensorflow2015-whitepaper} on its backend. \textit{med7} is used for extracting clinical related entities from clinical notes. All experiments experiments were performed on a computer with NVIDIA Tesla K80 GPU with 24GB of VRAM, 378 GB of ram and Intel Xeon E5 2683 processor. The full code of this work is available at \url{https://github.com/tanlab/ConvolutionMedicalNer}.

\renewcommand{\arraystretch}{1.5}
\setlength{\tabcolsep}{10pt}

\begin{table*}[!t]
\centering
\resizebox{\textwidth}{!}{\begin{tabular}{l c c c c c}
\hline
Task & Modal & Embedding & AUROC & AUPRC & F1\\ \hline
\multirow{3}{*}{\textbf{In-Hospital Mortality}} 

& Best Baseline & - & 86.42 $\pm$ 0.004 & 54.47 $\pm$ 0.007 & 46.60 $\pm$ 0.016\\
\cline{2-6}
&                  & Word2Vec & \textbf{87.55} $\pm$ 0.003 & \textbf{55.87} $\pm$ 0.008 & \textbf{47.23} $\pm$ 0.014\\
 & Proposed Model        &  FastText & 87.15 $\pm$ 0.002 & 55.68 $\pm$ 0.005 & 46.87 $\pm$ 0.015\\
 &                  & Concat & 86.98 $\pm$ 0.003 & 55.35 $\pm$ 0.008 & 46.38 $\pm$ 0.027\\
\hline
\hline

\multirow{3}{*}{\textbf{In-ICU Mortality}} 

& Best Baseline & - & 87.17 $\pm$ 0.002 & 48.47 $\pm$ 0.006 & 42.91 $\pm$ 0.014\\
\cline{2-6}
 &                  & Word2Vec & \textbf{88.35} $\pm$ 0.002 & \textbf{49.23} $\pm$ 0.008 & 43.02 $\pm$ 0.029\\
  & Proposed Model             &  FastText & 87.85 $\pm$ 0.001 & 48.78 $\pm$ 0.009 & \textbf{43.09} $\pm$ 0.026\\
  &                 &  Concat & 87.66 $\pm$ 0.002 & 48.74 $\pm$ 0.009 & 42.24 $\pm$ 0.027\\
\hline
\hline

\multirow{3}{*}{\textbf{LOS $>$ 3 Days}} 

& Best Baseline & - & 68.90 $\pm$ 0.002 & 61.88 $\pm$ 0.002 & 54.70 $\pm$ 0.009\\
\cline{2-6}
 &                  &  Word2Vec & 69.54 $\pm$ 0.002 & 62.68 $\pm$ 0.003 & 55.04 $\pm$ 0.012\\
  & Proposed Model         &  FastText & 69.61 $\pm$ 0.003 & 62.55 $\pm$ 0.003 & \textbf{55.87} $\pm$ 0.017\\
  &                 &  Concat & \textbf{69.93} $\pm$ 0.001 & \textbf{62.77} $\pm$ 0.002 & 55.82 $\pm$ 0.008\\
\hline
\hline

\multirow{3}{*}{\textbf{LOS $>$ 7 Days}} 

& Best Baseline & - & 71.63 $\pm$ 0.005 & 17.91 $\pm$ 0.006 & \textbf{2.33} $\pm$ 0.012\\
\cline{2-6}
 &                  &  Word2Vec &\textbf{ 72.55} $\pm$ 0.005 & \textbf{18.78} $\pm$ 0.006 & 1.58 $\pm$ 0.001\\
  & Proposed Model             &  FastText & 71.81 $\pm$ 0.004 & 18.01 $\pm$ 0.004 & 1.08 $\pm$ 0.008\\
  &                  &  Concat & 71.92 $\pm$ 0.007 & 18.25 $\pm$ 0.006 & 1.38 $\pm$ 0.009\\
\hline
\hline

\end{tabular}}
\caption{Proposed model performance comparison with best baseline model. We select the highest score for each metric and each clinical task from baseline methods.}
\label{tab:proposed-results}
\end{table*}

\subsection{Results}
\label{subsec:results}

\subsubsection{Baseline Modal Results}
\label{subsubsec:time-series-results}
We predict four different clinical tasks with the patient's first 24 hours ICU measurements and medical entities. Table~\ref{tab:baseline-scores} summarizes the overall performance of baseline methods. As seen from results, instead of strong results of time-series GRU model, multimodal approaches improve the performance, as expected. For in-hospital mortality prediction, we see an improvement of \%1.5 AUROC, \%2.5 AUPRC and \%4 F1 score compare to the time-series GRU modal. For other mortality prediction task, in-icu mortality, multimodal approach improve the performance around \%2 for AUROC and AUPRC and \%7 for F1 score. Multimodal approach also improves the performance of predictions tasks in LOS problem. Both in LOS $>$ 3 and LOS $>$ 7, all metrics are improved around \%1.5. For all experiments, time-series GRU modal only get better F1 score for LOS $>$ 7 problem.

\subsubsection{Proposed Modal Results}
\label{subsubsec:multimodal-results}

In this section, we compare the result of our proposed model against the best scores taken from baseline models. All results for the proposed model against best baseline scores are provided in Table~\ref{tab:proposed-results}. As shown in Table~\ref{tab:baseline-scores}, multimodal approach improves the performance of predictions tasks over the time-series, however we try to use medical entities more efficiently to improve the prediction of our models. Except the F1 score of LOS $>$ 7 clinical task, our proposed multimodal architecture robustly outperforms all other baseline models for each task. 

\section{Discussion}
\label{sec:discussion}

Table~\ref{tab:baseline-scores} shows that the use of medical entity features improve the prediction performance on all clinical tasks. As shown in Table~\ref{tab:baseline-scores}, multimodal baseline modals increase all metrics performance which indicates the benefit of using medical entities for predicting mortality and LOS. These experiments also provide an opportunity to compare the medical entity representation methods. Although there is no certain winner for all tasks, in the baseline models, the results show us for mortality prediction tasks, representing the medical entities with averaging method gives better results. For LOS prediction tasks, representing all medical entities together with Doc2Vec is also successful as averaging method. Furthermore, both scores on Table~\ref{tab:baseline-scores} and Table~\ref{tab:proposed-results} gives us a chance to compare the word embedding approaches. We do not observe a significant change in performance between word embedding techniques, however pretrained Word2Vec model generally achieves slightly higher scores (around \%0.5) than FastText and experimental concatenated embeddings. Apart from these experiments and comparisons, our main motivation is finding an efficient way to combine time-series features with medical entities. Even though both baseline multimodals improve the prediction results compared to timeseries baseline, to make better feature extraction on medical entities, we want to take the advantage of 1D CNN. In the literature, there have been several studies that use 1D CNN in NLP. We stack three 1D convolution operation to extract the features, and then apply 1D max pooling operation over the time-step to obtain a fixed-length vector. By analyzing the results between the proposed and baseline multimodals, we see that 1D CNN based multimodal approach give better results than the averaging and document based embedding methods. Addition to these trials, we also make experiments by using only medical entity features as another baseline. However, only medical entity baseline give poor results (around less than \%10 for all tasks) compared to the timeseries and multimodal, so we do not report these results.

\section{Conclusion}
\label{sec:conclusion}

Over the past decade, there has been increased attention to improve mortality and LOS prediction performance. Predicting any complications and saving patient's life is an important task for healthcare system which motivates us to work on mortality prediction. LOS is another important clinical problem to improve hospital performance and better healthcare resource utilisation. In this work, we present 1D-CNN based multimodal deep learning architecture that use time-series features and medical entities together and this model outperforms several baselines. Our proposed model performance gain over multimodal baselines is around \%1 - \%1.5 AUPRC, and the improvement over time-series baseline is around \%2.5 - \%3 AUPRC. We also make experiments to investigate the effect of different word embedding algorithms to solve our clinical problems and report the results. This work can be extended in multiple directions. First, we can involve more features associated with patient such as prescription data and diagnosis codes to improve the prediction performance. Second, using different word embedding especially transformer based techniques can be used for learning the entity representations. Another thing we may consider in the future is to use more advanced deep learning architectures with attention based will be useful for clinical tasks.

\bibliographystyle{unsrt}
\bibliography{references.bib}

\end{document}